\documentclass{article}

\usepackage{arxiv}

\usepackage[utf8]{inputenc} 
\usepackage[T1]{fontenc}    
\usepackage{hyperref}       
\usepackage{url}            
\usepackage{booktabs}       
\usepackage{amsmath,amssymb,amsfonts}       
\usepackage{nicefrac}       
\usepackage{microtype}      
\usepackage{lipsum}
\usepackage{graphicx}
\usepackage{algorithmic}
\usepackage{algorithm}

\title{An Incremental Construction of Deep Neuro Fuzzy System for Continual Learning of Non-stationary Data Streams \thanks{This paper has been published in IEEE Transactions on Fuzzy Systems.}}

\author{
  Mahardhika Pratama\\
  School of Computer Science and Engineering\\
  Nanyang Technological University\\
  Singapore\\
  \texttt{mpratama@ntu.edu.sg} \\
   \And
 Witold Pedrycz\\
  Department of Electrical and Computer Engineering\\
  University of Alberta\\
  Canada\\
  \texttt{wpedrycz@ualberta.ca} \\
  \And
  Geoffrey I. Webb\\
  Information Technology Faculty\\
  Monash University\\
  Australia\\
  \texttt{geoff.webb@monash.edu}
}

\begin{document}
\maketitle

\begin{abstract}
Existing fuzzy neural networks (FNNs) are mostly developed under a shallow network configuration having lower generalization power than those of deep structures. This paper proposes a novel self-organizing deep fuzzy neural network, namely deep evolving fuzzy neural networks (DEVFNN). Fuzzy rules can be automatically extracted from data streams or removed if they play limited role during their lifespan. The structure of the network can be deepened on demand by stacking additional layers using a drift detection method which not only detects the covariate drift, variations of input space, but also accurately identifies the real drift, dynamic changes of both feature space and target space. DEVFNN is developed under the stacked generalization principle via the feature augmentation concept where a recently developed algorithm, namely Generic Classifier (gClass), drives the hidden layer. It is equipped by an automatic feature selection method which controls activation and deactivation of input attributes to induce varying subsets of input features. A deep network simplification procedure is put forward using the concept of hidden layer merging to prevent uncontrollable growth of dimensionality of input space due to the nature of feature augmentation approach in building a deep network structure. DEVFNN works in the sample-wise fashion and is compatible for data stream applications. The efficacy of DEVFNN has been thoroughly evaluated using seven datasets with non-stationary properties under the prequential test-then-train protocol. It has been compared with four popular continual learning algorithms and its shallow counterpart where DEVFNN demonstrates improvement of classification accuracy. Moreover, it is also shown that the concept drift detection method is an effective tool to control the depth of network structure while the hidden layer merging scenario is capable of simplifying the network complexity of a deep network with negligible compromise of generalization performance.
\end{abstract}

\keywords{Deep Neural Networks \and Data Streams \and Online Learning \and Fuzzy Neural Network}

\section{Introduction}
Deep neural network (DNN) has gained tremendous success in many real-world problems because its deep network structure enables to learn complex feature representations \cite{DeepIOT}. Its structure is constructed by stacking multiple hidden layers or classifiers to produce a high level abstraction of input features which brings improvement of model's generalization. There exists a certain point where introduction of extra hidden nodes or models in a wide configuration has negligible effect toward enhancement of generalization power. The power of depth has been theoretically proven \cite{powerofdepth} with examples that there are simple functions in $d$-dimensional feature space that can be modelled with ease by a simple three-layer feedforward neural network but cannot be approximated by a two-layer feedforward neural network up to a certain accuracy level unless the number of hidden nodes is exponential in the dimension. Despite of these aforementioned working advantages, the success of DNN relies on its static network structure blindly selected by trial-error approaches or search methods \cite{NIPS2012_4824}. Although a very deep network architecture is capable of delivering satisfactory predictive accuracy, this approach incurs excessive computational cost. An over-complex DNN is also prone to the so-called vanishing gradients \cite{DeepExpandable} and diminishing feature reuse \cite{CirculantProjections}. In addition, it calls for a high number of training samples to ensure convergence of all network parameters.
	
To address the issue of a fixed and static model, model selection of DNN has attracted growing research interest. It aims to develop DNN with elastic structure featuring stochastic depth which can be adjusted to suit the complexity of given problems \cite{stochasticdepth}. In addition, a flexible structure paradigm also eases gradient computation and addresses the well-known issues: diminishing feature reuse and vanishing gradient. This approach starts from an over-complex network structure followed by complexity reduction scenario via dropout, bypass, highway \cite{CirculantProjections}, hedging \cite{OnlineDeepLearning}, merging \cite{Net2NetL}, regularizer \cite{LearningTheNumber}, etc. Another approach utilizes the idea of knowledge transfer or distillation \cite{distilling}. That is, the training process is carried out using a very deep network structure while deploying a shallow network structure during the testing phase. Nonetheless, this approach is not scalable for data stream applications because most of which are built upon iterative training process. Notwithstanding that \cite{OnlineDeepLearning} characterizes an online working principle, it starts its training process with a very deep network architecture and utilizes the hedging concept which opens direct link between hidden layer and output layer. The final output is determined from aggregation of each layer output. This strategy has strong relationship to the weighted voting concept.
	
The concept of DNN is introduced into fuzzy system in \cite{stackedDFNN,stackedDFNN2,stackedDFNN3} making use of the stacked generalization principle \cite{stackedgeneralization}. Two different architectures, namely random shift \cite{stackedDFNN} and feature augmentation \cite{stackedDFNN2,stackedDFNN3}, are adopted to create deep neuro fuzzy structure. It is also claimed that fuzzy rule interpretability is not compromised under these two structures because intermediate features still have the same physical meaning as original input attributes. Similar work is done in \cite{onlinedeepTSK} but the difference exists in the parameter learning aspect adopting the online learning procedure instead of the batch learning module. These works, however, rely on a fixed and static network configuration which calls for prior domain knowledge to determine its network architecture. In \cite{deepsemisupervised}, a deep fuzzy rule-based system is proposed for image classification. It is built upon a four-layered network structure where the first three layer consists of normalization layer, scaling layer and feature descriptor layer while the final layer is a fuzzy rule layer. Unlike \cite{stackedDFNN,stackedDFNN2,stackedDFNN3,OnlineDeepLearning}, this algorithm is capable of self-organizing its fuzzy rules in the fuzzy rule layer but the network structure still has a fixed depth (4 layer). Although the area of deep learning has grown at a high pace, the issue of data stream processing or continual learning remains an open issue in the existing deep learning literature.
	
A novel incremental DNN, namely Deep Evolving Fuzzy Neural Network (DEVFNN), is proposed in this paper. DEVFNN features a fully elastic structure where not only its fuzzy rule can be autonomously evolved but also the depth of network structure can be adapted in the fully automatic manner \cite{DSSCN}. This property is capable of handling dynamic variations of data streams but also delivering continuous improvement of predictive performance. The deep structure of DEVFNN is built upon the stacked generalization principle via the augmented feature space where each layer consists of a local learner and is inter-connected through augmentation of feature space \cite{stackedDFNN2,stackedDFNN3}. That is, the output of previous layer is fed as new input information to the next layer. A meta-cognitive Scaffolding learner, namely Generic Classifier (gCLass), is deployed as a local learner, the main driving force of hidden layer, because it not only has an open structure and works in the single-pass fashion but also answers two key issues: what-to-learn and when-to-learn \cite{gClass}. The what-to-learn module is driven by an active learning scenario which estimates contribution of data points and selects important samples for model updates while the when-to-learn module controls when the rule premise of multivariate Gaussian rule is updated. gClass is structured under a generalized Takagi Sugeno Kang (TSK) fuzzy system which incorporates the multivariate Gaussian function as the rule premise and the concept of functional link neural network (FLANN) \cite{FLANN} as the rule consequent. The multivariate Gaussian function generates non axis-parallel ellipsoidal clusters while the FLANN expands the degree of freedom (DoF) of the rule consequent via the up to second order Chebyshev series rectifying the mapping capability\cite{patra2002nonlinear}.
	
The major contribution of this paper is elaborated as follows:
\begin{itemize}
	\item \textit{Elastic Deep Neural Network Structure}: DEVFNN is structured by a deep stacked network architecture inspired by the stacked generalization principle. Unlike the original stacked generalization principle having two layers, DEVFNN’s network structure is capable of being very deep with the use of feature augmentation concept. This approach adopts the stacked deep fuzzy neural network concept in \cite{stackedDFNN2} where the feature space of the bottom layer to the top one is growing incorporating the outputs of previous layers as extra input information. DEVFNN differs itself from \cite{stackedDFNN2,stackedDFNN3,onlinedeepTSK} where it characterizes a fully flexible network structure targeted to address the requirement of continual learning \cite{DeepIOT}. This property is capable of expanding the depth of the DNN whenever the drift is identified to adapt to rapidly changing environments. The use of drift detection method for the layer growing mechanism generates different concepts across each layer supposed to induce continuous refinement of generalization power. The elastic characteristic of DEVFNN is borne out with the introduction of a hidden layer merging mechanism as a deep structure simplification approach which shrinks the depth of network structure on the fly. This mechanism focuses on redundant layers having high mutual information to be coalesced with minor cost of predictive accuracy.
	\item \textit{Dynamic Feature Space Paradigm}: an online feature selection scenario is integrated in the DEVFNN learning procedure and enables the use of different input combinations for each sample.  This scenario enables flexible activation and deactivation of input attributes across different layers which prevents exponential increase of input dimension due to the main drawback of feature augmentation approach. As with \cite{wang2014online,featurerelevance}, the feature selection process is carried out with binary weights (0 or 1) determined from the relevance of input features to the target concept. Such feature selection method opens likelihood of previously deactivated features to be active again whenever its relevance is substantiated with current data trend. Moreover, the dynamic feature space paradigm is realized using the concept of hidden layer merging method which functions as a complexity reduction approach. The use of hidden layer merging approach has minor compression loss because one layer can be completely represented by another hidden layer.
	\item \textit{An Evolving Base Building Unit}: DEVFNN is constructed from a collection of Generic Classifier (gClass) \cite{gClass} hierarchically connected in tandem. gClass functions as the underlying component of DEVFNN and operates in every layer of DEVFNN. gClass features an evolving and adaptive trait where its structural construction process is fully automated. That is, its fuzzy rules can be automatically generated and pruned on the fly. This property handles local drift better than a non-evolving base building unit because its network structure can expand on the fly. The prominent trait of gClass lies in the use of online active learning scenario as the what-to-learn part of metacognitive learner which supports reduction of training samples and labeling cost. This strategy differs from \cite{pensembleplus} since the sample selection process is undertaken in a decentralized manner and not in the main training process of DEVFNN. Moreover, the how-to-learn scenario is designed in accordance with the three learning pillars of Scaffolding theory \cite{Leanr++NSE}: fading, problematizing and complexity reduction which processes data streams more efficiently than conventional self-evolving FNNs due to additional learning modules: local forgetting mechanism, rule pruning and recall mechanism, etc.
	\item \textit{Real Drift Detection Approach}: a concept drift detection method is adopted to control the depth of network structure. This idea is confirmed by the fact that the hidden layer should produce intermediate representation which reveals hidden structure of data samples through multiple linear mapping. Based on the recent study \cite{deeplinear}, it is outlined from the hyperplane perspective that the number of response region has a direct correlation to model$'$s generalization power and DNN is more expressive than a shallow network simply because it has much higher number of response region. In realm of deep stacked network, we interpret response region as the amount of unique information a base building unit carries. In other words, the drift detection method paves a way to come up with a diverse collection of hidden layers or building units. Our drift detection method is a derivation of the Hoeffding bound based drift detection method in \cite{drift} but differs from the fact that the accuracy matrix which corresponds to prequential error is used in lieu of sample statistics \cite{Pesaranghader2018}. This modification targets detection of real drift which moves the shape of decision boundary. Another salient aspect of DEVFNN$'$s drift detector exists in the confidence level of Hoeffding$'$s bound which takes into account sample$'$s availability via an exponentially decreasing confidence level. The exponentially decreasing confidence parameter links the depth of network structure and sample’s availability. This strategy reflects the fact that the depth of DNN should be adjusted to the number of training samples as well-known from the deep learning literature. A shallow network is generally preferred for small datasets because it ensures fast model convergence whereas a deep structure is well-suited for large datasets.
	\item \textit{Adaptation of Voting Weight}: the dynamic weighting scenario is implemented in the adaptive voting mechanism of DEVFNN and generates unique decaying factors of every hidden layer. Our innovation in DEVFNN lies in the use of dynamic penalty and reward factor which enables the voting weight to rise and decline with different rates. It is inspired by the fact that the voting weight of a relevant building unit should decrease slowly when making misclassification whereas that of a poor building unit should increase slowly when returning correct prediction. This scenario improves a static decreasing factor as actualized in pENsemble \cite{pENsemble}, pENsemble+ \cite{pensembleplus}, DWM \cite{Kolter2003} which imposes too violent fluctuations of voting weights.
\end{itemize}
	
The novelty of this paper is summed up in five facets: 1) contributes methodology in building the structure of DNNs which to the best of our knowledge remains a challenging and open issue; 2) offers a DNN variant which can be applied for data stream processing; 3) puts forward online complexity reduction mechanism of DNNs based on the hidden layer merging strategy and the online feature selection method; 4) introduces a dynamic weighting strategy with the dynamic decaying factor. The efficacy of DEVFNN has been numerically validated using seven synthetic and real-world datasets possessing non-stationary characteristics under the prequential test-then-train approach - a standard evaluation procedure of data stream algorithms. DEVFNN has been also compared by its static version and other online learning algorithms where DEVFNN delivers more encouraging performance in term of accuracy and sample consumption than its counterparts while imposing comparable computational and memory burden. This paper is organized as follows: Section 2 outlines the network architecture of DEVFNN and its hidden layer, gClass; Section 3 elaborates the learning policy of DEVFNN encompassing the hidden layer growing strategy, the hidden layer merging strategy and the online feature selection strategy; Section 4 offers brief summary of gClass learning policy; Section 5 describes numerical study and comparison of DEVFNN; some concluding remarks are drawn in the last section of this paper.

\section{Problem Formulation}
DEVFNN is deployed in the continual learning environment to handle data streams $C_t=[C_1,C_2,...,C_T]$ which continuously arrives in a form of data batch in the $T$ time stamps. In practise, the number of time stamps is unknown and is possible to be infinite. The size of data batch is $C_t=[X_1,X_2,...,X_P]\in\Re^{P\times n}$ and is set to be equal-sized in every time stamp. Note that the size of data batches may vary as well across different time stamps. $n$ stands for the number of input attributes. In realm of online learning environments, it is impractical to assume the direct access to the true class label vector $Y_t=[y_1,y_2,...,y_P]\in\Re^{P\times m}$ and labelling process often draws some costs subject to the existence of ground truth or expert knowledge. $m$ here stands for the number of target classes. This fact confirms the suitability of prequential test-then-train scenario as evaluation procedure of the continual learning. The $1-0$ encoding scenario is applied to form a multi-output target matrix. For instance, if a data samples lies in a class $1$ and there exists in total three classes, the $1-0$ encoding scheme produces a target vector $[1,0,0]$, while a class 2 leads to $[0,1,0]$. The issue of labelling cost is tackled with the use of metacognitive learner as a driving force of hidden layer having the online active learning scenario as a part of the what-to-learn phase.
	
Data stream is inherent to the problem of concept drift where data batches do not follow static and predictable data distributions $P(Y|X)_t\neq P(Y|X)_{t-1}$ well-known to be in two typical types: real and covariate. The real drift is generally more challenging to handle than the covariate drift because variations of input data induce the shape of decision boundary usually resulting in classification error. This problem cannot be solved by inspecting the statistic of input data as proposed in \cite{drift}. One approach to capture the presence of concept drift is by constructing accuracy metric which summarizes predictive performance of a classifier. A decreasing trend of classifier$'$s accuracy is a strong indicator of concept drift \cite{Pesaranghader2018}. This issue is resolved here by the use of accuracy matrix which advances the idea of accuracy matrix in \cite{Pesaranghader2018} with an adaptive windowing scheme based on the cutting point concept. The presence of concept drift also calls for innovation in the construction of deep network structure because every layer is supposed to represent different concepts rather than different levels of data abstraction. DEVFNN handles this issue with direct connection of hidden layer to output layer under the deep stacked network structure with the feature augmentation approach in which it allows every layer to contribute toward the final classification decision. Specifically, the weighting voting scheme is implemented where the voting weight is derived from the reward and punishment approach with dynamic decaying rates. The dynamic decaying rate sets unique reward and penalty factors giving added flexibility according to relevance of every hidden layer.  On the other hand, the feature augmentation method in building deep network structure risks on the curse of dimensionality due to the expansion of input dimension as the network depth. The network simplification procedure is realized with the hidden layer merging procedure and the online feature selection scenario. Fig. 1 delineates the working principle of DEVFNN.

\section{Network Architecture of DEVFNN}
This section elaborates on the network architecture of DEVFNN including both the deep network structure and the hidden layer structure.
\begin{center}
	\begin{figure*}\label{Architecture of DSSCN1}
		\begin{centering}
			\includegraphics[scale=0.8]{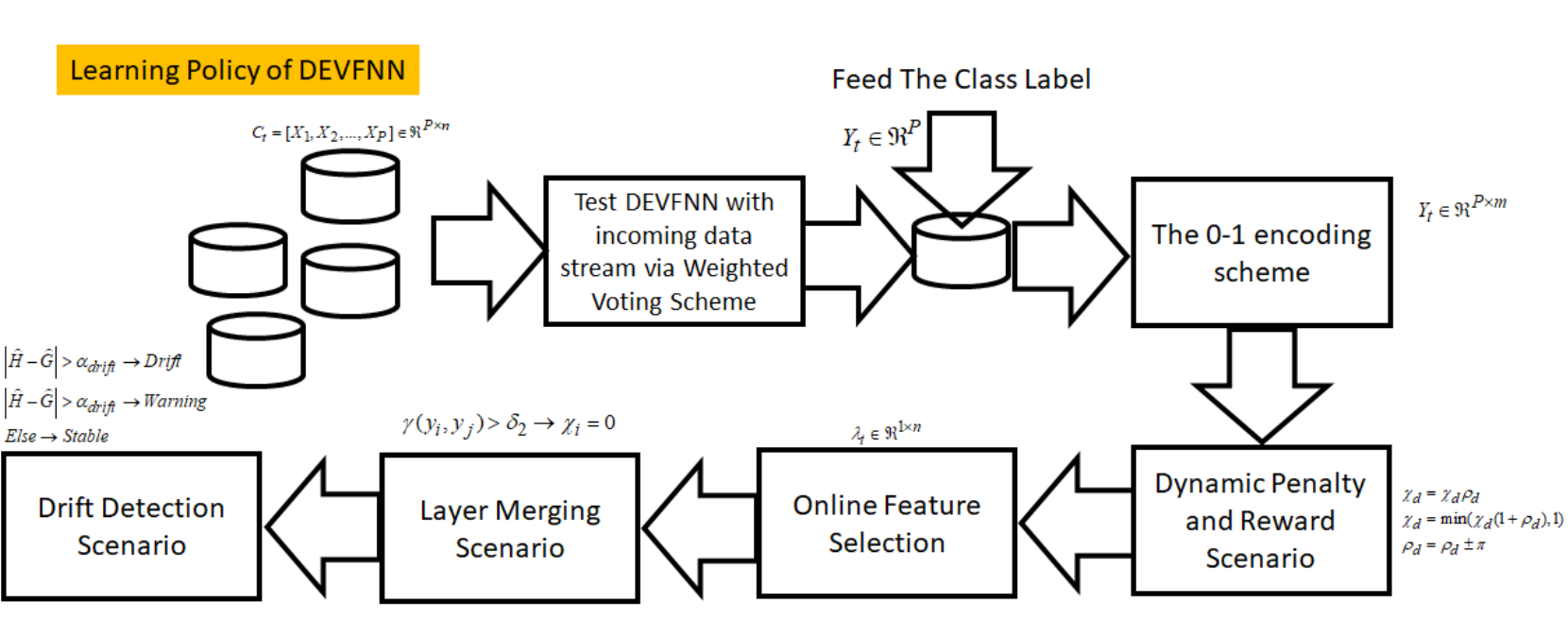}
			\par\end{centering}
		\caption{Learning Architecture of DEVFNN}
	\end{figure*}
	\par\end{center}

\subsection{Architecture of Hidden Layer}
The hidden layer of DEVFNN is constructed by Generic Classifier (gClass) \cite{gClass} working cooperatively in tandem to produce the final output. gClass realizes the generalized TSK fuzzy inference system procedure where the multivariate Gaussian function with non diagonal covariance matrix is used in the rule layer while the consequent layer adopts the concept of expansion block of the FLANN via a nonlinear mapping of the up-to second order Chebyshev series \cite{patra2002nonlinear}. The multivariate Gaussian rule generates better input space partition than those classical rules using the dot product t-norm operator because it enables rotation of ellipsoidal cluster \cite{pratama2014genefis}. This trait allows a reduction of fuzzy rule demand especially when data samples are not spanned in the main axes. The use of FLANN idea in the consequent layer \cite{FLANN} aims to improve the approximation power because it has a higher DoF than the first order TSK rule consequent. Several nonlinear mapping functions such as polynomial power function, trigonometric function, etc. can be applied as the expansion block but the up-to second order Chebyshev function is utilized here because it incurs less free parameters than trigonometric function but exhibits better approximation power than other polynomial functions of the same order \cite{PRVFLN}.
	
The fuzzy rule of gClass is formally expressed as follows:

$\Re_i$: \textbf{IF}    $X_k$ is $N_i(X_k;C_i,A_{i}^{-1})$ \textbf{Then} $\tilde{y_i}=\Phi W_i=W_{0,i}T_0+W_{1,i}T_1(x_1)+W_{2,i}T_2(x_2)+...+W_{2n-1,i}T_1(x_n)+W_{2n,i}T_2(x_n)$

where $X_k\in\Re^{1\times n}$ denotes the input vector at the $k-th$ time instant and $N_i(X_k;C_i,A_{i}^{-1})$ stands for the multivariate Gaussian function which corresponds to the rule premise of $i-th$ fuzzy rule. $\tilde{y_i}$ labels the rule consequent of the $i-th$ rule constructed by the up to second order Chebyshev polynomial expansion where $T_0(.),T_1(.),T_2(.)$ refer to the zero, first and second order Chebyshev polynomial expansion respectively. The multivariate Gaussian function generates the firing strength of $i-th$ fuzzy rule which reveals its degree of compatibility as follows:
\begin{equation}
\phi_i=exp(X-C_i)A_i^{-1}(X-C_i);
\end{equation}
where $C_i\in\Re^{1\times n}$ denotes the center of multivariate Gaussian function while $A_i^{-1}\in\Re^{n \times n}$ stands for the non-diagonal inverse covariance matrix while $R,n$ are respectively the number of fuzzy rules and input features. The non-diagonal covariance matrix makes possible to describe inter-relations among input variables which vanishes in the case of diagonal covariance matrix and steers the orientation of ellipsoidal cluster. On the other hand, the rule consequent of gClass is built upon the non-linear mapping of the up-to second order Chebyshev series defined as $\tilde{y_i}=\Phi W_i$ where $\Phi\in\Re^{1 \times (2n+1)}$ is the output of functional expansion block while $W_i\in\Re^{(2n+1)\times 1}$ is the weight vector of the $i-th$ rule. $\Phi$ is produced by the up-to second order Chebyshev function expressed as follows:
\begin{equation}
T_{n+1}(x_j)=2x_jT_n(x_j)-T_{n-1}(x_j)
\end{equation}
Since the Chebyshev series is expanded up to the second order, $T_0(x_j)=1,T_1(x_j)=x_j,T_2(x_j)=2x_j^{2}-1$. Suppose that a predictive task is navigated by two input attributes, the functional expansion block $\Phi$ is crafted as follows:
\begin{eqnarray}
\Phi=[T_0(x_j),T_1(x_1),T_2(x_j),T_1(x_2),T_2(x_2)]\\
\Phi=[1,x_1,2x_1^{2}-1,x_2,2x_2^{2}-1]
\end{eqnarray}
The concept of FLANN in the framework of TSK fuzzy system improves approximation capability of zero or first order TSK rule consequent because it enhances the degree of freedom of rule consequent, although it does draw around extra $n$ network parameters. Under the MIMO structure, the rule consequent incurs $(2n+1)\times R \times m$ parameters while the rule premise of gClass imposes $(n\times n)\times R+(n\times R)$.

The output of gClass is inferred using the weighted average operation of the rule firing strength and the output of functional expansion block as follows
\begin{equation}
y_o=\frac{\sum_i^R\phi_i\tilde{y_i}}{\sum_i^R\phi_i}=\frac{\sum_i^R exp(X-C_i)A_i^{-1}(X-C_i)^{T}\tilde{y_i}}{\sum_i^R exp(X-C_i)A_i^{-1}(X-C_i)^{T}}
\end{equation}
gClass is formed in the MIMO architecture possessing the class-specific rule consequent \cite{pratama2015pclass,classifierarchitecture} addressing the issue of class overlapping better than the popular one-versus-rest architecture because each class is handled separately. Because gClass generates multiple output, the final classification decision is obtained by applying the maximum operation to find the most confident prediction as follows:
\begin{equation}
y=\max_{1 \leq o \leq m} y_o
\end{equation}
where $m$ is the number of target classes. The MIMO structure is also more stable than the direct regression approach hampered by the class shifting issue \cite{pratama2015pclass,classifierarchitecture}. This problem is caused by the smooth transition of the TSK fuzzy system which cannot cope with a dramatic class change.

\begin{center}
	\begin{figure*}\label{Architecture of DSSCN}
		\begin{centering}
			\includegraphics[scale=0.8]{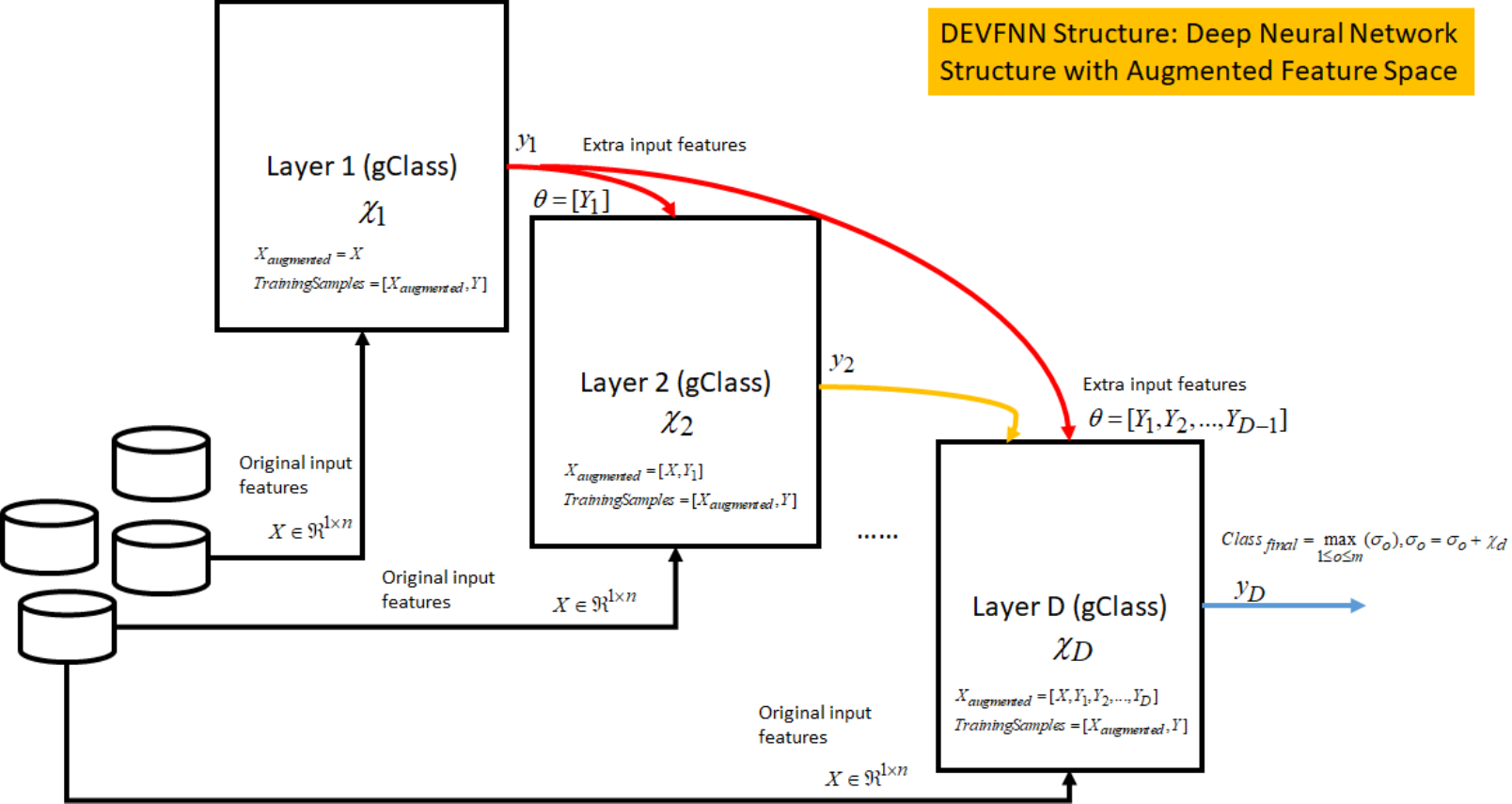}
			\par\end{centering}
		\caption{Deep Stacked Network Structure with Feature Augmentation}
	\end{figure*}
	\par\end{center}
	
\subsection{Deep Stacked Network}
DEVFNN is realized in the deep stacked network structure where each hidden layer is connected in series. Hidden layers except the first layer accept predictive outputs of previous layers as input attributes plus original input patterns \cite{stackedDFNN2}. This trait retains the physical meaning of original input variables which often loses in conventional DNN architectures due to multiple nonlinear mappings. This aspect also implies three exists a growth of input space across each layer. That is, the deeper the hidden layer position the higher the input dimension is, since it is supposed to receive all predictive outputs of preceding layers as extra input features. This structure is inspired by \cite{stackedDFNN2,stackedDFNN3} where the concept of feature augmentation is introduced in building a DNN architecture. Our approach extends this approach with the introduction of structural learning process which fully automates the construction of deep stacked network via online analysis of dynamic data streams. Moreover, this approach is equipped by the online feature selection and hidden layer merging scenarios which dynamically compress the input space of each layer. This mechanism replaces the random feature selection of the original work in \cite{stackedDFNN2,stackedDFNN3}. Our approach also operates in the one-pass learning scenario being compatible with online learning setting whereas \cite{stackedDFNN2,stackedDFNN3} still applies an offline working scenario. The network architecture of DEVFNN is shown in Fig. 2.

DEVFNN works in the chunk-by-chunk basis where each chunk is discarded once learned and the chunk size may vary in different time stamps. Suppose that a data chunk with the size of $P$ $C_t=(X_k,Y_k)_{k=1,...,P}$ is received at the $k-th$ time stamp while DEVFNN is composed of $D$ hidden layers built upon gClass, the first hidden layer receives the original input representation with no feature augmentation while the $2-nd$ hidden layer is presented with the following input attributes:
\begin{equation}
\widehat{X}_k^{2}=[X_k,\widehat{Y}_k^{1}]
\end{equation}
where $\widehat{X}_k^{2}\in\Re^{P\times(n+m)}$ consists of the input feature vector and the predictive output of the first layer. Additional $m$ input attributes are incurred because gClass is structured under the MIMO architecture deploying a class-specific rule consequent. For $d-th$ hidden layer, its input features are expressed as $\widehat{X}_k^{d}=[X_k,\widehat{Y}_k^{1},...,\widehat{Y}_k^{d-1}]_{k=1,...,P}\in\Re^{P\times(n+(m\times(d-1)))}$. This situation also implies that dimensionality of input space linearly increases as the depth of network structure. This issue is coped with the self-organizing property of DEVFNN featuring a controlled growth of network depth by means of the drift detection method which assures the right network complexity of a given problem. In addition, the hidden layer merging and online feature selection mechanisms further contribute to compression of input dimension to handle possible curse of dimensionality.

The deep stacked network through feature augmentation is consistent to the stacked generalization principle \cite{stackedgeneralization} which is capable of achieving an improved generalization performance since the manifold structure of the original feature space is constantly opened. In addition, each hidden layer functions as a discriminative layer producing predicted class label. The predicted class label delivers an effective avenue to open the manifold structure of the original input space to the next hidden layer supposed to improve generalization power.

The end output of DEVFNN is produced by the weighted voting scheme where the voting weight of hidden layer increases and decreases dynamically in respect to their predictive accuracy. This strategy reflects the fact that each hidden layer is constructed with different concepts navigated by the drift detection scenario. That is, a hidden layer is rewarded when returning correct prediction by enhancing its voting weight while being punished by lowering its voting weight when incurring misclassification. It reduces the influence of irrelevant classifier where it is supposed to carry low voting power due to the penalty as a result of misclassification \cite{pENsemble}. It is worth noting that the predictive performance of hidden layers is examined against the prequential error when true class labels are gathered. The prequential test-then-train scenario is simulated in our numerical study. It enables creation of real data stream environment where data streams are predicted first and used for model updates once the true class label becomes available. This scenario is more plausible to capture real data stream environments in practise in which true class labels often cannot be immediately fed by the operator. Furthermore, DEVFNN is composed of independent building blocks interconnected with the feature augmentation approach meaning that each hidden layer is trained independently with the same cost function and outputs the same target variables. The final output of DEVFNN is formalized as follows:
\begin{equation}
Class_{final}=\max_{1 \leq o \leq m} \sigma_{o}
\end{equation}
where $\sigma_o$ denotes the accumulated weight of a target class resulted from predicted class label of hidden layers $\sigma_{o}=\chi_{d}^{class}+\sigma_{o}$. $\chi_{d}^{o}$ stands for the voting weight of $d-th$ hidden layer predicting $o-th$ class label. This procedure aims to extract the most relevant subsets of hidden layer to the $C_t$ data batch. That is, each hidden layer is set to carry different concepts and is triggered when the closest concept comes into the picture. In addition, the tuning phase is restricted to the $win$ hidden layer in order to induce a stable concept for every hidden layer. That is, irrelevant hidden layers are frozen from the tuning phase when they represent different concepts. The $win$ hidden layer can be determined from its voting weight where a high voting weight indicates strong relevance of a hidden layer to the current concept.

\section{Learning Policy of DEVFNN}
This section elaborates the learning policy of DEVFNN in the deep structure level. An overview of DEVFNN learning procedure is outlined in Algorithm 1. DEVFNN initiates its learning process with the adaptation of voting weight with dynamic decaying factors. The tuning of voting weight results in a unique voting weight of each hidden layer with respect to their relevance to the current concept measured from their prequential error. This method can be also interpreted as a soft structural simplification method since it is capable of ruling out hidden layers in the training process by suppressing their voting weights to a small value. The training procedure continues with the online feature selection procedure which aims to cope with the curse of dimensionality due to the growing feature space bottleneck of the deep stacked network structure. The feature selection is performed based on the idea of feature relevance examining the sensitivity of input variables in respect to the target classes. The hidden layer merging mechanism is put forward to capture redundant hidden layers and reduces the depth of network structure by coalescing highly correlated layers. Furthermore, the drift detection mechanism is applied to deepen the deep stacked network if changing system dynamics are observed in data streams. The drift detection module is designed using the concept of Hoeffding bound \cite{drift,Pesaranghader2018} which determines a conflict level to flag a drift. There exist three conditions resulted from the drift detector: stable, warning and drift. The drift phase leads to addition of a hidden layer, while the warning phase accumulates data samples into a buffer later used to create a new hidden layer if a drift is detected. The stable phase simply updates the winning hidden layer determined from its voting weight. In what follows, we elaborate on the working principle of DEVFNN:

\subsection{Adaptation of Voting Weight}
The voting weight is dynamically adjusted by a unique decaying factor for each layer $\rho_d \in [0,1]$ which plays a major role to adapt to the concept drift. A low value of $\rho_d$ leads to slow adaptation to rapidly changing conditions but works very well in the presence of gradual or incremental drift where the drift is not too obvious initially. A high value of $\rho_d$ adapts to sudden drift more timely than the low value but compromises the stability in the gradual drift where data samples are drawn from the mixture between two distributions during the transition period. Considering this issue, $\rho_d$ is not kept fixed rather it continuously adjusted to reflect the learning performance of a hidden layer. The decaying factor $\rho_d$ is tuned using a step size, $\pi$, as follows:
\begin{equation}
\rho_d=\rho_d \pm \pi;
\end{equation}
This mechanism underpins a flexible voting scenario where the decaying factor mirrors hidden layer compatibility to existing data streams. In other words, the voting weight of hidden layers augments and diminishes with different intensities. It is also evident that the voting weight of a strong hidden layer should be confirmed whereas the influence of poor building units should be minimized in the voting phase. That is, $\rho_d=\rho_d+\pi$ occurs if a hidden layer returns a correct prequential prediction whereas $\rho_d=\rho_d-\pi$ takes place if a sample leads to a prequential error. The step size $\pi$ sets the rate of change where the higher the value increases its sensitivity to the hidden layer performance.

The penalty and reward scenario is undertaken by increasing and decreasing the voting weight of a hidden layer. A penalty is imposed if a wrong prediction is produced by hidden layer as follows:
\begin{equation}
\chi_d=\chi_d*\rho_d;
\end{equation}
On the other hand, the reward scenario is performed if correct prediction is returned as follows:
\begin{equation}
\chi_d=min(\chi_d(1+\rho_d),1)\label{reward}
\end{equation}
the reward scenario also functions to handle the cyclic drift because it reactivates inactive hidden layer with a minor voting weight. Because it is also observed from (\ref{reward}) and the range of the decaying factor $\rho_d$ is $[0,1]$, the voting weight is bounded in the range of $[0,1]$. Compared to similar approaches in \cite{pENsemble,pensembleplus}, this penalty and reward scenario emphasizes rewards to a strong hidden layer while discouraging a penalty to such layers whereas a poor hidden layer should be penalized with a high intensity while receiving little reward when returning correct prediction. This strategy is done since every hidden layer has direct connection to output layer outputting its own predicted class label. The final predicted class label should take into account relevance of each hidden layer reflected from its prequential error. Furthermore, this mechanism also aligns with the use of drift detection method as an avenue to deepen the depth of network structure since the drift detection method evolves different concepts of each hidden layer.

\subsection{Online Feature Weighting Mechanism}
DEVFNN implements the feature weighting strategy based on the concept of feature relevance. That is, a relevant feature is defined as that showing high sensitivity to the target concept whereas low sensitivity feature signifies poor input attributes which should be assigned a low weight to reduce its influence to the final predictive outcomes \cite{lughofer2015generalized}. There exist several avenues to check the sensitivity of input attributes in respect to the target concept: Fisher Separability Criterion \cite{onlinefeatureweighting}, statistical contribution \cite{pratama2014genefis}, etc. The correlation measure is considered as the most plausible strategy here because it reveals the mutual information of input and target features \cite{featurerelevance} which signals the presence of changing system dynamics. The correlation between two variables, $x_1,x_2$, can be estimated using the Pearson correlation index (PCI) as follows:

\begin{equation}
\zeta(x_1,x_2)=\frac{cov(x_1,x_2)}{\sqrt{var(x_1,x_2)}}
\end{equation}
where $cov(x_1,x_2),var(x_1,x_2)$ respectively stand for covariance and variance of $x_1,x_2$ which can be calculated in recursive manners. Although the PCI can be directly integrated into the feature weighting scope without any transformation because the highest correlation is attained when it returns either $-1$ or $1$, the PCI method is sensitive to the rotation and translation of data samples \cite{mitra2002unsupervised}.  The mutual information compression index (MICI) method \cite{mitra2002unsupervised} is applied to achieve trade-off between accuracy of correlation measure and computational simplicity. The MICI method works by estimating the amount of information loss if one of the two variables is discarded. It is expressed as follows:
\begin{align}
\gamma(x_{1},x_{2})=\frac{1}{2}(vr_1+vr_2-\nonumber\\\sqrt{vr_1^{2}+vr_2^{2}-4vr_1vr_2(1-\zeta(x_{1},x_{2})^{2})})
\end{align}
where $\zeta{(x_1,x_2)}$ denotes the Pearson correlation index of two variables and $vr_1,vr_2$ represent $var(x_1),var(x_2)$, respectively. Unlike the PCI method where -1 and 1 signify the maximum correlation, the maximum similarity is attained at $\gamma(x_{1},x_{2})=0$. This method is also insensitive to rotation and translation of data points\cite{mitra2002unsupervised}. Once the correlation of the $j-th$ input variable and all $m$ target classes are calculated, the score of the $j-th$ input feature is defined by its relevance to all $m-th$ target classes. This aspect is realized by taking average correlation across $m$ target classes as follows:
\begin{equation}
Score_j=mean_{o=1,..,m}\gamma(x_j,y_o)
\end{equation}
where $\gamma(x_j,y_o)$ denotes the maximum information compression index between the $j-th$ input feature and the $o-th$ target class. The use of average operator is to assign equal importance of each target class and to embrace the fact that an input feature is highly needed to only identify one target class. This strategy adapts to the characteristic of gClass formed in the MIMO architecture \cite{classifierarchitecture}.

The feature selection mechanism is carried out by assigning binary weights $\lambda_j$, either 0 or 1, which changes on demand in every time stamp. This strategy is applied to induce flexible activation and deactivation of input variables during the whole course of training process which avoids loss of information due to complete forgetting of particular input information. An input feature is switched off by assigning 0 weight if its score falls above a given threshold $\delta_1$ meaning that it shows low relationship to any target classes as follows:
\begin{equation}
Score_j>=\delta_1
\end{equation}
where $\delta_1\in[0,1]$ stands for the predefined threshold. The higher the value of this threshold the less aggressive the hidden layer merging is performed and vice versa. The feature selection process is carried out by setting the input weight to zero with likelihood being set to one again in the future whenever the input attribute becomes relevant again. Moreover, the input weighting strategy is committed in the centralized manner where the similarity of all input attributes are analyzed at once. That is, all input attributes are put together. The online feature weighting scenario only analyses the $n$ dimensional original feature space rather than the $n+m\times(D-1)$ dimensional augmented feature space. Extra input feature is subject to the hidden layer merging mechanism which studies the redundancy level of $D$ hidden layers.

\subsection{Hidden Layer Merging Mechanism}
DEVFNN implements the hidden layer merging mechanism to cope with the redundancy across different base building units. This scenario is realized by analyzing the correlation of the outputs of different layers \cite{pensembleplus}. From manifold learning viewpoint, a redundant layer containing similar concept is expected not to inform salient structure of the given problem because it does not open manifold of learning problem to unique representation - at least already covered by previous layers. Suppose that the MICI method is applied to explore the correlation of two hidden layers and $\gamma(y_i,y_j),i\neq j$, the hidden layer merging condition is formulated as follows:
\begin{equation}
\gamma(y_i,y_j)<\delta_2
\end{equation}
where $\delta_2\in[0,1]$ is a user-defined threshold. This threshold is linearly proportional to the maximum correlation index where the lower the value the less merging process is undertaken. The merging procedure is carried out by setting its voting weight as zero, thus it is considered as a "don$'$t care" input attribute of the next layers. This strategy expedites the model updates because redundant layers can be bypassed without being revisited for both inference and training procedures. The hard pruning mechanism is not implemented in the merging process because it causes a reduction of input dimension which undermines the stability of next layers unless a retraining mechanism from scratch is carried out. It causes dimensional reduction of the output covariance matrix of next layers. This strategy is deemed similar to the dropout scenario \cite{dropout} in the deep learning literature but the weight setting is analyzed from the similarity analysis rather than purely probabilistic approach. The dominance of two hidden layers is simply determined from its voting weight. The voting weight is deemed a proper indicator of hidden layer dominance because it is derived from a dynamic penalty and reward scenario with unique and adaptive decaying factors. The redundancy-based approach such as merging scenario is more stable than relevance-based approach because information of one layer can be perfectly represented by another layer. In addition, parameters of two hidden layers are not fused because similarity of two hidden layers is observed in the output level rather than fuzzy rule level.

\subsection{Hidden Layer Growing Mechanism}
DEVFNN realizes an evolving deep stacked network structure which is capable of introducing new hidden layers on demand. This strategy aims to embrace changing training patterns of data streams and to enhance generalization performance by increasing the level of abstraction of training data. This is done by utilizing a drift detection module which vets the status of data streams whether the concept change is present. Notwithstanding that the idea of adding a new component to handle concept drift is well-established in the literature as presented in the ensemble learning literature, each ensemble member has no interaction at all to their neighbors.

The drift detection scenario makes use of the Hoeffding$'$s bound drift detection mechanism proposed in \cite{drift} where the evaluation window is determined from the switching point rather than a fixed window size. Nevertheless, the original method is developed based on the increase of population mean which ignores the change of data distribution in the target space. In other words, it is only capable of detecting covariate drift. Instead of looking at the increase of population mean, the error index is applied. This modification is based on the fact that the change of feature space does necessarily induce the concept drift in the target domain $P(Y|X)_{k}\neq P(Y|X)_{k-1}$. This concept is implemented by constructing a binary accuracy vector where $0$ elements presents correct prediction while $1$ elements are inserted for false predictions. This scenario is inspired by the fast Hoeffding drift detection method (FDDM) \cite{Pesaranghader2018} but our approach differs from \cite{drift} in which the window size is set fully adaptive according to the switching point. Moreover, each data sample is treated with equal importance with the absence of any weights which detects sudden drift rapidly although it is rather inaccurate to pick up gradual drift \cite{drift}. The advantage of the Hoeffding$'$s method is free of normal data distribution assumption - too restrictive in many applications. Moreover, it is statistically sound because a Hoeffding bound corresponds to a particular confidence level.

Assuming that $P$ denotes the chunk size, the data chunk is partitioned into three groups $F\in\Re^{P\times(n+m)},G\in\Re^{cut\times(m+n)},H\in\Re^{(P-cut)\times(m+n)}$ where $cut$ is the switching point. $F,G,H$ records the error index instead of the original data points in which only two values, namely $0$ or $1$, is present - $0$ for true prediction, $1$ for false prediction. Note that $cut$ is elicited by evaluating data samples - the first sample up to the switching sample. Each data partition $F,G,H$ is assigned with the error bounds $\epsilon_F,\epsilon_G,\epsilon_H$ calculated as follows:
\begin{equation}
\epsilon_{F,G,H}=(b-a)\sqrt{\frac{size}{2 (size*cut)}\ln({\frac{1}{\alpha}})}
\end{equation}
where $size$ denotes the size of data partition and $\alpha$ labels the significance level of Hoeffding's bound. $a,b$ denote the maximum and minimum values in the data partition. The significance level has a clear statistical interpretation because it corresponds to the confidence level of Hoeffding's bound $1-\alpha$. In realm of DNN, the model$'$s complexity must consider the availability of training samples to ensure parameter convergence especially in the context of continual learning situation where a retraining process over a number of epochs is prohibited. A shallow model is generally preferred over a deep model to handle a small dataset.  Considering the aspect of sample$'$s availability, a dynamic significance level is put forward where the significance level exponentially rises as the number of time stamps with a limit. A limit is required here to avoid loss of detection accuracy because the significance level is inversely proportional to the confidence level:
\begin{equation}
\alpha_{D}=min(1-e^{\frac{-k}{T}},\alpha_{min}^{D}),\alpha_{W}=min(1-e^{\frac{-k}{T}},\alpha_{min}^{W}) \label{drift}
\end{equation}
where $k,T$ respectively denote the number of time stamps seen thus far and the total number of time stamps while $\alpha_{min}^{D},\alpha_{min}^{W}$ stand for the minimum significance level  of the drift and warning phases. The minimum significance level has to be capped at 0.1 to induce above 90\% confidence level.

Once the confidence level is calculated, the next step is to find the switching point, $cut$, indicating the horizon of the drift detection problem or the time window in which a drift is likely to be present. The switching point is found if the following condition is met as follows:
\begin{equation}
\hat{F}+\epsilon_{F}\leq \hat{G}+\epsilon_{G}\label{switching point}
\end{equation}
where $\hat{F},\hat{G},\hat{H}$ denote the statistics of the three data partitions. It is observed that the switching point targets a transition point between two concepts where the statistic of $G$ is larger than $H$. The switching point portrays a time index where a drift starts to come into picture. It is worth noting that the statistics of the data partition $\hat{G}$ is expected to be constant or to reduce during the stable phase. The drift phase, therefore, refers to the opposite case where the empirical mean of the accuracy vector $\hat{G}$ increases. The condition (\ref{switching point}) aims to find the cutting point $cut$ where the accuracy vector is no longer in the decreasing trend. Once $cut$ is located, the three error index vectors, $F,G,H$, can be formed.

Our drift detector returns two conditions: warning and drift tailored from the two significance levels $\alpha_{warning},\alpha_{drift}$. The two significance levels $\alpha_{warning},\alpha_{drift}$ can be set to the confidence level of the drift detection. The smaller the value of the significance level implies the more accurate the drift detection method delivers. The warning and drift conditions are signaled if two following situations are come across as follows:
\begin{equation}
|\hat{H}-\hat{G}|> \alpha_{drift}
\end{equation}
\begin{equation}
|\hat{H}-\hat{G}|> \alpha_{warning}
\end{equation}
These two conditions present a case where the null hypothesis $H_{0}:E[G]\leq E[H]$ is rejected. The warning condition is meant to capture the gradual drift. It pinpoints a situation where a drift is not obvious enough to be declared or next instances are called for to confirm a drift. No action is taken during the warning phase, only data samples of the warning condition are accumulated in the data matrix $\phi=[X_{warning},Y_{warning}]$. Once the drift condition is satisfied, the structure of deep network is deepened by appending a new hidden layer stacked at the top level. The new hidden layer is created using the incoming data chunk and the accumulated data samples in the warning phase $\Phi=[X_{new},Y_{new};\phi]$.

The opposite case or the normal case portrays the alternative $H_{1}:E[G]> E[H]$. That is, the stable phase only activates an adjustment of the winning hidden layer to improve the generalization power of DEVFNN. The winning hidden layer is selected for the adaptation scenario instead of all hidden layers to expedite the model update. Moreover, DEVFNN adopts the concept of different-depth network structure which opens the room for each layer to produce the end-output of DEVFNN because each layer is trained to solve the same optimization problem. Because the dynamic voting mechanism is implemented, the winning hidden layer is simply selected based on its voting weight.

\section{gClass learning procedure}
This section provides brief recap of gClass working principle \cite{gClass} constructed under three learning pillars of meta-cognitive learning: what-to-learn, how-to-learn and when-to-learn. The what-to-learn component functions as the sample selection module implemented under the online active learning scenario, while the when-to-learn component sets the sample update condition. The how-to-learn module realizes a self-adaptive learning principle developed under the framework of Scaffolding theory \cite{Leanr++NSE}: problematizing, fading and complexity reduction. This procedure encompasses rule generation, pruning, forgetting and tuning mechanisms. A  flowchart of gClass learning policy is placed in the supplemental document.
\begin{algorithm}
	\caption{Learning Policy of DEVFNN}
	\label{devfnn}
	\begin{algorithmic}[1]
		\STATE \textit{Input}: $(X_n,Y_n)\in\Re^{P\times(n+m)}$, $\pi$, $\delta_1$, $\delta_2$, $Vig$ and $k_{prune}$
		\STATE \textit{Output}: $Class_{final}$ final predicted Class
		\STATE \textbf{Testing phase}:
		\STATE \textit{Predict}: The class label of the current data batch. 
		\STATE \textbf{Training phase}:
		\STATE {Step 1: Tuning of $\chi$}:
		\FOR {$k = 1$ to $P$}
		\FOR {$d=1$ to $D$}
		\IF{($\hat{y}_k^d\neq y_k$)}
		\STATE \textit{Execute}: $\rho_d=\rho_d-\pi$ and $chi_d=\chi_d*\rho_d$
		\STATE \textit{Update}: the accuracy matrix $Acc(k)=1$
		\ELSE
		\STATE \textit{Execute}: $\rho_d=\rho_d+\pi$, $\chi_d=min(\chi_d(1+\rho_d),1)$
		\STATE \textit{Update}: the accuracy matrix $Acc(k)=0$
		\ENDIF
		\ENDFOR
		\ENDFOR\\
		\STATE {Step 2: Online Feature Selection Mechanism}:
		\FOR {$j = 1$ to $n$}
		\FOR {$o=1$ to $m$}
		\STATE \textit{Calculate}: the input target correlation (12)
		\IF{($score>\delta_1$)}
		\STATE \textit{Set}: the input weight to $\lambda_j=0$
		\ENDIF
		\ENDFOR
		\ENDFOR\\
		\STATE {Step 3: Hidden Layer Merging Mechanism }:
		\FOR {$d_1 = 1$ to $D$}
		\FOR {$d_2=1$ to $D$}
		\STATE \textit{Calculate}: the correlation coefficient (14) \IF{($\gamma(y_{d_1},y_{d_2}) <\delta_1, \forall d_1 \neq d_2$)}
		\STATE \textit{Set}: the voting weight of weaker layer to 0
		\ENDIF
		\ENDFOR
		\ENDFOR\\
		\STATE {Step 4: Drift Detection Mechanism}:
		\FOR {$k = 1$ to $P$}
		\STATE \textit{Construct}: $F=Acc,G=Acc(1:k)$
		\STATE \textit{Calculate}: $\epsilon_F,\epsilon_G, \hat{G},\hat{H}$ (15)
		\IF{($\hat{G}+\epsilon_G \geq \hat{F}+\epsilon_F$)}
		\STATE \textit{Set}: $cut=k$
		\STATE \textit{Construct}: $H=Acc(P-cut+1:cut)$ and $\hat{H}$
		\ENDIF
		\ENDFOR\\
		\STATE \textit{Calculate}: the drift and warning level $\epsilon_{drift},\epsilon_{warning}$
		\IF{($|\hat{G}-\hat{F}|\geq \epsilon_{drift}$)}
		\STATE \textit{Deepen}: the network structure using $(X_n,Y_n)\in \Re^{P\times(n+m)}$ and $\phi$
		\ELSIF{($|\hat{G}-\hat{F}|\neq \epsilon_{drift} \text{and} |\hat{G}-\hat{F}|\geq \epsilon_{warning}$)}
		\STATE \textit{Create}: $\phi$
		\ELSE
		\STATE \textit{Update}: The winning layer using $(X_n,Y_n)\in \Re^{P\times(n+m)}$
		\ENDIF
	\end{algorithmic}
\end{algorithm}

\begin{itemize}
	\item \textit{What-To-Learn Phase - Online Active Learning Scenario}: The online active learning scenario of gClass is driven by the extension of extended conflict ignorance (ECI) principle \cite{Lughofer2012} which relies on two sample contribution measures. The first concept is developed from the idea of extended recursive density estimation (ERDE) applied as the rule growing scenario in \cite{angelov2004approach}. Our approach distinguishes itself from \cite{angelov2004approach} in which the RDE concept is modified for the multivariate Gaussian rule and integrates the sample weighting concept overcoming the outlier$'$s bottleneck. Unlike the rule growing concept finding salient samples as those having maximum and minimum densities, the sample of interest for deletion purpose is redundant samples defined as those violating the maximum and minimum conditions. The second approach is designed using the distance-to-boundary concept. A classifier is said to be confident if it safely classifies a sample to one of classes or is far from decision boundary. The sample-to-boundary concept is set as a ratio of the first and second dominant classes examined by the classifier$'$s outputs. An uncertain sample is indicated if the ratio returns a value around 0.5 whereas a high ratio signifies a confident case.
	\item \textit{How-To-Learn - Sample Learning Strategy}: Once a sample is accepted by the online active learning scenario, it is passed to the how-to-learn scenario evolving parameter and structure of gClass. The problematizing part concerns on the issue of concept drift handling and the complexity reduction part relieves the problem$'$s complexity while the fading part is devised for reduction of model$'$s complexity. The three components are integrated in a single dedicated learning process and executed in the one-pass learning scenario.
	
	The problematizing phase consists of two learning modules: the rule growing and forgetting scenarios. The rule growing phase of gClass adopts the same criteria of pClass \cite{pratama2015pclass} in which the three rule growing conditions, namely data quality (DQ), datum significance (DS), volume check, are consolidated to pinpoint an ideal observation to expand the rule base size. The data quality method is a derivation of the RDE method in \cite{angelov2004approach} involving the sample weighting concept. Moreover, it is tailored to accommodate the multivariate Gaussian rule. The DS concept estimates the statistical contribution of a data sample whether it deserves to be a candidate of new rule \cite{pratama2014panfis}, while the volume check is integrated to prevent the over-sized rule which undermines model$'$s generalization. The DS concept extends the concept of neuron significance \cite{GGAPRBF} to be compatible with the non axis-parallel ellipsoidal rule while the volume check examines the volume of winning rule and a new rule is introduced given that its volume exceeds a pre-specified limit. Once the three rule growing conditions are satisfied, a new fuzzy rule is initialized using the class overlapping situation. The class overlapping method arranges the three initialization strategies in respect to spatial proximity of a data sample to inter-class and intra-class data samples. This strategy is carried our using the quality per class method which studies relationship of current sample to target classes. The rule forgetting scenario is carried out in the local mode which deploys unique forgetting levels of each rule \cite{localdrift}. It is derived using the local DQ (LDQ) method performing recursive local density estimation. The gradient of LDQ method for each rule is calculated and used to define the forgetting level in the rule premise and consequent. Moreover, the rule consequent tuning scenario is driven by the fuzzily weighted generalized recursive least square (FWGRLS) method inspired by the work of generalized recursive least square method (GRLS) method \cite{GRLS} putting forward the weight decay term in the cost function of RLS method. This method can be also perceived as a derivation of FWRLS method in \cite{angelov2004approach} in which it improves the tuning scenario with the weight decay term to improve model$'$s generalization.
	
	The fading phase lowers structural complexity of gClass to avoid the overfitting problem and to expedite the runtime. The fading phase of gClass is crafted under the same strategy as pClass \cite{pratama2015pclass} where two rule pruning strategy, namely extended rule significance (ERS) and potential+ (P+) methods, are put forward. The ERS method shares the same principle as the DS method where it approximates the statistical contribution of fuzzy rules. This approach can be also seen as an estimator of expected outputs of gClass under uniform distribution. The P+ method adopts the concept of rule potential \cite{angelov2004approach} but converts this method to perform the rule pruning task. Unlike the ERS method capturing superfluous rules playing little role during their lifespan, the P+ method discovers obsolete rule, no longer relevant to represent the current data distribution. By extension, the P+ method is also applied in the rule recall scenario to cope with the cyclic drift. That is, obsolete rules are only deactivated with possibility to be reactivated again in the future if it becomes relevant again.
	
	gClass implements the online feature weighting mechanism based on the fisher seperability criteria (FSC) in the empirical feature space using the kernel concept as adopted in pClass \cite{pratama2015pclass} as a part of complexity reduction method. Nevertheless, the online feature weighting scenario of gClass is switched into a sleep mode because DEVFNN is already equipped by an online feature selection and layer merging module in the top level.
	
	\item \textit{When-To-Learn - Sample Reserved Strategy}: The when-to-learn strategy of gClass is built upon the standard sample reserved strategy of the meta-cognitive learner \cite{das2015evolving}. Our approach, however, differs from that \cite{das2015evolving} where the sample learning condition is designed under different criteria. The sample reserved strategy incorporates a condition for the tuning scenario of rule premise and data samples violating the rule growing and tuning procedures are accumulated in a data buffer reserved for the future training process. Our approach simply exempts those samples and exploit them merely for the rule consequent adaptation scenario because such sample quickly become outdated in rapidly changing environments. In addition, This procedure is designated to reduce the computational cost. A flowchart of DEVFNN learning policy is attached in the supplemental document.
\end{itemize}

\section{Numerical Study}
This section discusses experimental study of DEVFNN in seven popular real-world and synthetic data stream problems: electricity-pricing, weather, SEA, hyperplane, SUSY, kddCUP and indoor RFID localization problem from our own project. DEVFNN is compared against  prominent continual learning algorithms in the literature: PNN \cite{progressiveneuralnetworks}, HAT \cite{hardattention}, DEN \cite{DeepExpandable}, DSSCN \cite{DSSCN}, staticDEVFNN. PNN, HAT and DEN are popular continual learning algorithms in the deep learning literature also designed to prevent the catastrophic forgetting problems. They feature self-evolution of hidden nodes but still adopt the static network depth. DSSCN represents a deep algorithm having stochastic depth. The key difference with our approach lies in the concept of random shift to form deep stacked network structure rather than the feature augmentation approach. Comparison with staticDEVFNN is important to demonstrate the efficacy of flexible structure. That is, the hidden layer expansion and merging modules are switched off in the static DEVFNN. The MATLAB codes of DEVFNN are made publicly available in \footnote{$https://bit.ly/2ZfnN5y$}. Because of the page limit, only SEA problem is discussed in the paper while the remainder of numerical study is outlined in the supplemental document. Nonetheless, Table 1 displays numerical results of all problems. Our numerical study follows the prequential test-then-train protocol. That is, a dataset is divided into a number of equal-sized data batches. Each data batch is streamed to DEVFNN in which it is used to first test DEVFNN's generalization performance before being used for the training process. This scenario aims to simulate real data stream environment \cite{GamaDataStream} where numerical evaluation is independently performed per data batch. The numerical results in Table 1 are reported as the average of numerical results across all data batches. Five evaluation criteria are used here: Classification Rate (CR), Fuzzy Rule (FR), Precision (P), Recall (R), Hidden Layer (HL). 
\begin{table}[t]
		\caption{Numerical results of consolidated algorithms}
		\begin{centering}
		\label{tab:Weather}
		\scalebox{0.8}{
		\begin{tabular}{llrrrrr}
				\toprule
				&  & CR  & FR  & P  & R & HL\tabularnewline
				\midrule 
				SEA  & DEVFNN & 91.7$\pm$ 5 & 4.36$\pm$ 1.08 & 0.95 & 0.92 & 1.01$\pm$ 0.1\tabularnewline
				 & SDEVFNN & 91.1$\pm$ 6.05 & 9 & 0.9 & 0.95 & 3\tabularnewline
				 & DSSCN & 91.5$\pm$ 4.09 & 10.29$\pm$ 4.09 & 0.92 & 0.95 & 1\tabularnewline 
				& PNN & 83.2$\pm$ 6.3 & 33 & 0.85 & 0.73 & 3\tabularnewline
				& DEN & 62.9$\pm$ 7.7 & 6 & 0.63 & 0.99 & 1\tabularnewline
				& HAT & 74.6$\pm$ 10.1 & 10 & 0.79 & 0.86 & 2\tabularnewline
				\midrule 
				Hyperplane & DEVFNN  & 91.57$\pm$ 1.76 & 4.73$\pm$ 1.36 & 0.92 & 0.91 & 1.58$\pm$ 0.5\tabularnewline
				 & SDEVFNN  & 55.52$\pm$ 14.2 & 4 & 0.53 & 0.99 & 3\tabularnewline
				& DSSCN  & 91.25$\pm$ 1.79 & 5.47$\pm$ 3.23 & 0.91 & 0.91 & 1\tabularnewline
				& PNN & 85.55$\pm$ 5.83 & 42 & 0.11 & 0.1 & 3\tabularnewline
				& DEN & 91.14$\pm$ 3.86 & 8 & 0.91 & 0.92 & 1\tabularnewline
				& HAT & 76.18$\pm$ 7.82 & 12 & 0.69 & 0.96 & 2\tabularnewline
				\midrule 
				Weather & DEVFNN  & 80$\pm$ 3.75 & 4.7$\pm$ 1.6 & 0.9 & 0.83 & 1.58$\pm$ 0.5\tabularnewline
				 & SDEVFNN  & 68$\pm$ 1.23 & 24 & 0.51 & 0.97 & 3\tabularnewline
				& DSSCN  & 80$\pm$ 2 & 5.08$\pm$ 2.9 & 0.71 & 0.59 & 10.62$\pm$ 3.88\tabularnewline
				& PNN & 68.7$\pm$ 1.18 & 78 & 0.3 & 0.93 & 3\tabularnewline
				& DEN & 68.46$\pm$ 5.16 & 10 & 0 & 0 & 1\tabularnewline
				& HAT & 67.36$\pm$ 8.27 & 20 & 0.008 & 0.027 & 2\tabularnewline
				\midrule 
				Electricity Pricing & DEVFNN  & 70$\pm$ 13 & 4.24$\pm$ 2.25 & 0.75 & 0.78 & 4.24$\pm$ 2.25\tabularnewline
				 & SDEVFNN  & 70$\pm$ 13 & 24 & 0.56 & 0.9 & 3\tabularnewline
				& DSSCN  & 68$\pm$ 13 & 3.58$\pm$ 1.45 & 0.72 & 0.62 & 3.58$\pm$ 1.45\tabularnewline
				& PNN & 59$\pm$ 3.4 & 78 & 0.35 & 0.66 & 3\tabularnewline
				& DEN & 57$\pm$ 8.6 & 16 & 0.08 & 0.16 & 1\tabularnewline
				& HAT & 57$\pm$ 8.4 & 20 & 0.1 & 0.15 & 2\tabularnewline
				\midrule 
				SUSY & DEVFNN  & 78$\pm$ 1.9 & 12.4$\pm$ 5.4 & 0.87 & 0.75 & 1$\pm$ 0.04\tabularnewline
				 & SDEVFNN  & 77$\pm$ 2.1 & 54 & 0.75 & 0.85 & 3\tabularnewline
				 & DSSCN  & 77$\pm$ 2 & 30.72$\pm$ 11.2 & 0.75 & 0.87 & 1\tabularnewline
				 & PNN & 72$\pm$ 3 & 30 & 0.75 & 0.56 & 3\tabularnewline
				& DEN & 64$\pm$ 10 & 10 & 0.76 & 0.92 & 1\tabularnewline
				& HAT & 72$\pm$ 3.4 & 20 & 0.63 & 0.95 & 2\tabularnewline
				\midrule 
				RFID & DEVFNN  & 92$\pm$ 11 & 2 & NA & NA & 1\tabularnewline
				 & SDEVFNN  & 90.7$\pm$ 8.3 & 9 & NA & NA & 3\tabularnewline
				 & DSSCN  & 91$\pm$ 14.6 & 10.2$\pm$ 5.2 & NA & NA & 10.18$\pm$ 5.2\tabularnewline
				 & PNN & 67$\pm$ 6.6 & 30 & NA & NA & 3\tabularnewline
				& DEN & 49$\pm$ 6.7 & 8 & NA & NA & 1\tabularnewline
				& HAT & 44.2$\pm$ 10.8 & 20 & NA & NA & 2\tabularnewline
				\midrule 
				KDD Cup 10\% & DEVFNN  & 99$\pm$ 0.48 & 1 & 0.99 & 0.99 & 1\tabularnewline
				 & SDEVFNN  & 99$\pm$ 0.005 & 123 & 0.99 & 0.99 & 3\tabularnewline
				 & DSSCN  & 93$\pm$ 10 & 1 & 0.99 & 0.99 & 1\tabularnewline
				 & PNN & 99$\pm$ 0.11 & 375 & 0.84 & 0.92 & 3\tabularnewline
				& DEN & 98$\pm$ 1.28 & 20 & 0.96 & 0.99 & 1\tabularnewline
				& HAT & 99$\pm$ 1.24 & 60 & 0.99 & 0.99 & 2\tabularnewline
				\bottomrule 
		\end{tabular}}
		\par\end{centering}
		\centering{} CR: classification rate, FR: fuzzy rule, P: precission, R: recall, HL: number of hidden layer, SDEVFNN: Static-DEVFNN 
	\end{table}

\subsection{SEA Problem}
The SEA problem is one of the most prominent problems in the data stream area where the underlying goal is to categorize data samples into two classes based on the summation of two input attributes \cite{SEA}. That is, a class 1 is returned if the condition $f_1+f_2<\theta$ while the opposite case $f_1+f_2>\theta$ indicates a class 2. The concept drift is induced by shifting the class threshold three times $\theta=4 \rightarrow 7 \rightarrow 4 \rightarrow 7$. This shift results in the abrupt drift and furthermore the cyclic drift because the class boundary is changed in the recurring fashion. The SEA problem is built upon three input attributes where the third input attribute is merely a noise. We make use of the modified SEA problem in \cite{DitzlerImbalanced} which incorporates 5 to 25\% minority class proportion. Data points are generated from the range of $[0,10]$ and the concept drift takes place in the target domain due to the drifting class boundary. The trace of classification rate, hidden layer and training sample are depicted in Fig. 4(a)-(d) of supplemental documents. Numerical results are reported in Table 1 along with numerical results of other problems.

The advantage of DEVFNN is reported in Table 1 where it outperforms other algorithms in terms of classification rates, precision and recall. This result highlights the efficacy of a deep stacked network structure compared to the static depth network in improving the generalization power. The use of a deep stacked network structure allows continuous refinement of predictive power where the output of preceding layer is fed as extra input information of current layer supposed to guide the predictive error toward zeros. The main bottleneck of the deep stacked network architecture through the feature augmentation approach lies in the linear increase of input dimensionality as the number of hidden layer. Notwithstanding that the hidden layer merging mechanism is integrated in DEVFNN and is supposed to lower the input dimension due to reduction of extra input attributes, the soft dimensionality reduction is applied by zeroing the voting weight of a hidden layer and the hidden layer is excluded from any training and inference processes. This strategy is to prevent instability issue due to the discontinuity of the training process which imposes a retraining process from scratch. Another complexity reduction method is implemented in terms of a dynamic voting weight scenario which overrides the influence of hidden layers in the final classification decision. This scenario generates unique reward and penalty weights giving heavy reward to a good hidden layer whereas strong penalty is imposed to that of poor hidden layers.

Because of the page limit, our numerical results of other datasets are presented in the supplemental document. A brief summary of our numerical results is as follows: 1) the decentralized online active learning strategy in the layer level is less efficient than the centralistic variant because it is classifier-dependent; 2) the proposal of real drift detection scenario is more timely than the covariate drift detection scenario because it reacts when the classifier$'$s performance is compromised; 3) the dynamic voting scenario with dynamic decaying factors lead to more stable voting scenario because the effect of penalty and reward can be controlled in respect to the classifier$'$s performance; 4) The application self-evolving depth enhances learning performance of the static depth because it grows the network on demand considering variation and availability of data samples.

\section{Conclusions}
A novel deep fuzzy neural network, namely dynamic evolving fuzzy neural network (DEVFNN), is proposed for mining evolving and dynamic data streams in the lifelong fashion. DEVFNN is constructed with a deep stacked network structure via the feature space augmentation concept where a hidden layer receives original input features plus the output of preceding layers as input features. This strategy generates continuous refinement of predictive power across a number of hidden layers. DEVFNN features an autonomous working principle where its structure and parameters are self-configured on the fly from data streams. A drift detection method is developed based on the principle of FDDM but it integrates an adaptive windowing scheme using the idea of cutting point. The drift detection mechanism is not only designed to monitor the dynamic of input space, covariate drift, but also to identify the nature of output space, real drift. Furthermore, DEVFNN is equipped by a hidden layer merging mechanism which measures correlation between two hidden layers and combines two redundant hidden layer. This module plays a key role in the deep stacked network structure via the feature augmentation concept to address uncontrollable increase of input dimension in rapidly changing conditions. DEVFNN also incorporates the online feature weighting method which assigns a crisp weight to input features in respect to their relevance to the target concept. The dynmic voting concept is introduced with the underlying notion "unique penalty and reward intensity" examined by the relevance of hidden layers.

DEVFNN is created with a local learner as a hidden layer, termed gClass, interconnected in tandem. gClass characterizes the meta-cognitive learning approach having three learning phases: what-to-learn, how-to-learn, when-to-learn. The what-to-learn and when-to-learn schemes provide added flexibility to gClass putting forward the online active learning scenario and the sample reserved strategy while the how-to-learn scheme is devised according to the Scaffolding theory. The online active learning method selects relevant samples to be labeled and to train a model which increases learning efficiency and mitigates the overfitting risk. The sample reserved strategy sets conditions for rule premise update while the scaffolding theory is followed to enhance the learning performance with several learning modules tailored according to the problematizing, fading and complexity reduction concepts of the Scaffolding theory.

The efficacy of DEVFNN has been numerically validated using seven prominent data stream problems in the literature where it produces more accurate classification rates, precision and recall than other benchmarked algorithms while incurring minor increase of computational and memory demand. It is found that the depth of network structure possesses a linear correlation with a generalization power given that every hidden layer is properly initialized and trained while the stochastic depth property improves learning performance compared to the static depth. Moreover, the dynamic adjustment of voting weights makes possible to adapt the voting weight with the dynamic adjustment factor which dynamically augments and shrinks according to its prequential error. It is perceived that the merging scenario dampens the network complexity without compromising classification accuracy. The concept drift detection method is applied to grow the hidden layer of network structure which sets flexibility for direct access of hidden layer to output layer. This implies that the final output is inferred by a combination of every layer output. In other words, DEVFNN actualizes a different-depth network paradigm where each level puts forward unique aspects of data streams. Our future work will investigate different approaches for self-generating the hidden layer of deep fuzzy neural network because it is admitted that the use of a concept drift detection method replaces the multiple nonlinear mapping of one concept generating a high level feature description with diverse concepts per layer.

\bibliographystyle{unsrt}  
\bibliography{references}  


\end{document}